\documentclass[letterpaper, 10 pt, conference]{ieeeconf}

\IEEEoverridecommandlockouts
\usepackage{cite}
\usepackage{algorithm}
\usepackage{algorithmic}
\usepackage[thinlines]{easytable}
\usepackage[pagewise]{lineno}
\usepackage{amsmath}
\DeclareMathOperator*{\argmax}{arg\,max}

\usepackage{graphicx}
\graphicspath{{./figures/}}

\title{\LARGE \bf
Trajectory Planning for Autonomous Vehicles Using Hierarchical Reinforcement Learning
}

\author{Kaleb Ben Naveed$^{1}$,  Zhiqian Qiao$^{2}$ and John M. Dolan$^{3}$
\thanks{$^{1}$Student of Electronic and Information Engineering, The Hong Kong Polytechnic University, Hong Kong, China. {\tt\small kaleb-ben.naveed@connect.polyu.hk}}%
\thanks{$^{2}$Electrical and Computer Engineering,  Carnegie Mellon University}
\thanks{$^{3}$ The Robotics Institute, Carnegie Mellon University}%
}        

\begin{document}

\maketitle
\thispagestyle{empty}
\pagestyle{empty}

\begin{abstract}

Planning safe trajectories under uncertain and dynamic conditions makes the autonomous driving problem significantly complex. Current sampling-based methods such as Rapidly Exploring Random Trees (RRTs) are not ideal for this problem because of the high computational cost. Supervised learning methods such as Imitation Learning lack generalization and safety guarantees. To address these problems and in order to ensure a robust framework, we propose a Hierarchical Reinforcement Learning (HRL) structure combined with a Proportional–Integral–Derivative (PID) controller for trajectory planning. HRL helps divide the task of autonomous vehicle driving into sub-goals and supports the network to learn policies for both high-level options and low-level trajectory planner choices. The introduction of sub-goals decreases convergence time and enables the policies learned to be reused for other scenarios. In addition, the proposed planner is made robust by guaranteeing smooth trajectories and by handling the noisy perception system of the ego-car. The PID controller is used for tracking the waypoints, which ensures smooth trajectories and reduces jerk. The problem of incomplete observations is handled by using a Long-Short-Term-Memory (LSTM) layer in the network. Results from the high-fidelity CARLA simulator indicate that the proposed method reduces convergence time, generates smoother trajectories, and is able to handle dynamic surroundings and noisy observations.

\end{abstract}


\section{INTRODUCTION}

Planning safe trajectories for autonomous vehicles is a challenging problem. In reality, this problem is particularly difﬁcult because of the maneuver planning complexities, stochastic surroundings, and the incomplete observations coming from the noisy perception system of the car. While performing trajectory planning, the autonomous vehicle has to plan different maneuvers, which might include lane following, waiting, changing lanes, and traversing intersections. 

The existing methods for trajectory planning either rely on traditional classical planners or machine learning methods. Some of the state-of-the-art traditional planners include sampling-based planners, such as Rapidly Exploring Random Trees (RRTs) and lattice planners. RRTs randomly build a space-ﬁlling tree and are well suited for an environment with obstacles and differential constraints [1]. However, their computational complexity increases 
\begin{figure}[htbp]
    \centering
    \includegraphics[width= 0.9\linewidth]{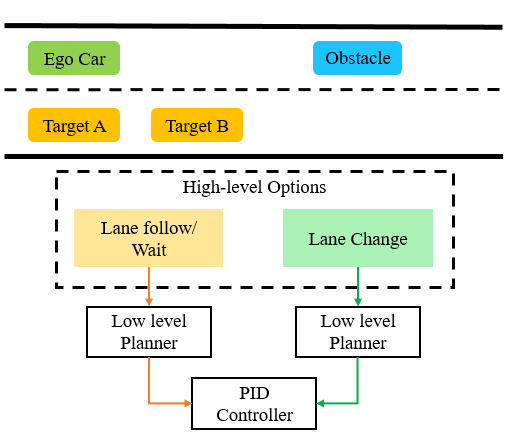}
    \caption{The Hierarchical structure has two high-level options and a low-level trajectory planner for each high-level option. Proportional-Integral-Derivative (PID) controller is used to follow the planned trajectory. The urban scenario is considered where ego-car has to either follow the lane or wait for the target-lane to be clear to perform a lane change maneuver.}
    \label{fig:Waiting}
\end{figure}
as the environment becomes more complex [2]. Moreover, the trajectories generated are often not smooth for the ego-car to follow [3]. On the other hand, lattice planners are great at generating feasible paths and incorporating constraints; however, they might create incomplete graphs, which lead to curvature discontinuity [2]. The other way of planning trajectories is through machine learning methods. The supervised learning method called Imitation Learning [4] has shown some promising results. However, this method might not generalize well to complex conditions and does not guarantee stability and an optimal solution [5]. 

An alternative approach to the classical planners and supervised learning methods is Reinforcement Learning (RL) [6]. The RL framework works on the principle of maximizing reward for a particular action at a given state. Existing RL works have shown promising results for an ego-car to learn policies for multiple scenarios. However, traditional RL methods for autonomous driving are less sample-efficient and less stable, especially for tasks with multiple sub-goals. In comparison to traditional RL, Hierarchical Reinforcement Learning (HRL) [7] allows a model to learn the policies for multiple sub-goals, which allows the policies learned to be reused for any other scenario. Furthermore, HRL has shown a faster convergence rate, which decreases training time for the model to learn an optimal policy. In this paper, we propose a HRL structure for trajectory planning. We choose the urban scenario of lane change shown in Figure 1 and give our structure two high-level options: a Lane follow/Wait and a Lane change option.  For each high-level option, there is a separate low-level trajectory planner. A detailed description of the low-level planner for each high-level option can be found in the Methodology section. The main contributions of the proposed HRL-based trajectory planner include:
\begin{itemize}
\item Planning safe and smooth trajectories: We use a PID controller for tracking trajectories instead of directly using the low-level actions: throttle, steer, and brake;
\item Dealing with noisy observations: We use  an LSTM layer in both networks, a high-level options network and a low-level planner network, to help the model learn from a sequence of observations, which compensates for missing or erroneous observations and also helps the ego-car to learn in more dynamic surroundings.
\end{itemize}

\section{RELATED WORK}

\subsection{Hierarchical Reinforcement Learning }

By extending the framework of RL, [7] proposed the idea of hierarchical Deep Q-Network (DQN), in which the action-value functions were integrated to operate at two levels of abstraction, and learned a policy over meta-goals and low-level actions. [8] proposed the concept of Hierarchical Q-Learning called MAXQ and showed that it produces better results than Q-Learning alone. In order to make the framework of hierarchical Q-Learning more robust, [9] combined the R-MAX algorithm with MAXQ. This amalgamation added a safe exploration dimension of the model-based approach into the MAXQ. In the autonomous vehicle’s domain, [10] used the three-layer HRL structure for options, low-level actions, and Q-network for the decision making problem for traversing intersections. They showed faster convergence rate and improved results compared to non-hierarchical approaches. In order to further improve the sample efficiency and the reward structure, [11] proposed the state-attention model, hybrid reward mechanism, and hierarchical prioritized experience replay. All these extensions improved sample efficiency and yielded better results.

\subsection{Work on Trajectory Planning and Prediction}

Existing work on trajectory planning includes both classical traditional planners and planners using machine learning principles. One of the state-of-the-art sampling-based planners used in autonomous driving planning is RRTs [1]. RRTs generate trajectories by constructing a tree-like structure through the space. RRTs are proven good for environments with obstacles, but might not converge to the optimal solution. [12] addressed this problem by introducing RRT*, which showed optimal convergence and shorter routes. Another approach introduced was lattice planners [2]. These planners are able to incorporate constraints and produce feasible paths. An alternative approach to classical planners includes supervised learning, which can be divided into imitation learning and trajectory prediction. [13] used a Deep Imitation Learning framework to learn a driving policy for the urban scenarios through offline learning. They also added a safety controller module which increased the safety while testing. 

For trajectory prediction, [14] proposed a system called UrbanFlow which consisted of a complete pipeline from collecting raw data to the final processing of trajectories. They used the UrbanFlow pipeline for trajectory prediction based on human drivers’ driving behavior, which allowed the ego-car to make better decisions. Another approach that has produced significant results in trajectory prediction is Inverse Reinforcement Learning (IRL). IRL works on the principle of extracting the reward structure by observing an optimal trajectory of an expert agent. By building on IRL approach, [15] proposed a framework for predicting off-road vehicle trajectories by integrating kinematics and environment to recover the reward structure.   

\subsection{Work on Lane Change }
Previous work on lane change includes both the use of classical control methods and learning algorithms. [16] proposed a solution to the lane change problem given that there was no road infrastructure support. They guided the ego-car to perform lane change by connecting the ego-lane to the target-lane using a virtual road curvature through the bicycle model. They also fed the steering angle into the bicycle model to estimate the lateral position to the target-lane. [17] proposed a RL-based methodology for the lane change situation. They combined the decision making of when to perform lane change through DQN with the Pure Pursuit Tracking algorithm for trajectory generation and tracking. This proposed structure combined machine learning techniques with classical control methods. Another work using learning methods introduced a Deep RL-based method to handle the ego-car's speed and lane change decisions. They used DQN to train the policy for speed and lane change decisions [18].

\subsection{Our Contribution}

In this paper, we make the RL framework for trajectory planning more robust. The proposed HRL framework for trajectory planning gives the ego-car two high-level options and three low-level planner choices for each high-level option. Each low-level planner selects waypoints based on the selected high-level option and the ego-car's state information, which is mainly its speed profile and distance to other vehicles.  The HRL modular structure ensures reusability for the policies learned and has shown a faster convergence rate. We build our model on the works of [10][11] but instead of directly using throttle, steer, and brake, we use a PID controller for tracking waypoints selected by the low-level planner. Trajectory tracking through PID resulted in smooth and stable maneuvers, which reduced jerk. Furthermore, in order to mitigate the problem of incomplete or inaccurate observations and dynamic environments, we added LSTM layers in the hierarchical network structure. 

\begin{figure}[htbp]
    \centering
    \includegraphics[width= 1.0\linewidth]{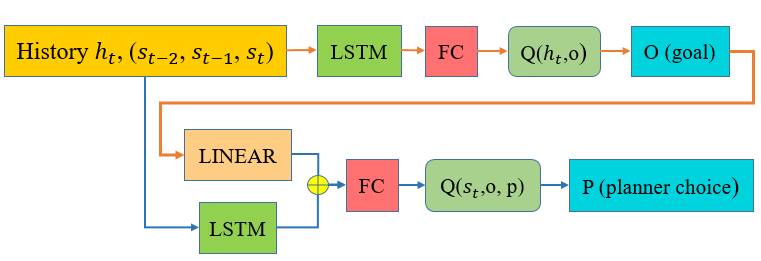}\caption{The Hierarchical RL network consists of a high-level Options Network and low-level Planner Network. In both networks, a Long Short Term Memory (LSTM) layer is used with a fixed number of previous steps $n_{steps} = 3$ for the history vector input. The $Tanh$ activation function is used in the LSTM layer. The output of the Options network, which is the goal, is concatenated with the history vector for the input to the planner network. A fully connected (FC) layer is used as a last layer for each network. Within the FC, a $ReLU$ activation function is used except for the last layer, which uses a $Linear$ activation function in order to generate both the network's final values.}
    \label{fig:Waiting}
\end{figure}

\section{PRELIMINARIES}

\subsection{Double Deep Q-Learning} Double Q-Learning is an extension of the state-of-the-art Deep Q-Learning algorithm. The Q-Learning algorithm in RL is used to find an optimal action-selection policy $\pi$ using a Q function which is used to maximize the action-value function $Q^*(s,a)$. 

Deep Q-Learning uses neural networks to update network parameter $\theta$ through minimizing the loss function between predicted action-value $Q$ and the target action-value $Y^Q$.
Deep Q-Learning uses the same values to select and evaluate an action, which results in overestimation.

Double Deep Q-Learning solves the problem of overestimation by revising the target action updates from another target $Q'$ network with different weights, which is shown in Equation 1.

\begin{equation} 
\label{eqn_example} 
Y_t^{Q} = R_{t+1} + \gamma  Q(S_{t+1},\argmax\limits_{a} Q(S_{t+1},a|\theta_{t}) |\theta_t^{'})
\end{equation}

\subsection{Hierarchical Reinforcement Learning}

HRL learns a policy at multiple levels as meta-controller $Q^1$ generates the sub-goal $g$ for the following steps and a controller $Q^2$ outputs the action $a$ based on the sub-goal selected until the next sub-goal is generated by the meta-controller.

\begin{equation} 
\label{eqn_example} 
Y_{t}^{Q^1} = \sum\limits_{t^{'}=t+1}^{t+1+N} R_{t^{'}} + \gamma \max\limits_{g}(S_{t+1+N}, g|\theta_{t}^1)
\end{equation}

\begin{equation} 
\label{eqn_example} 
Y_{t}^{Q^2} = R_{t+1} + \gamma \max\limits_{a}Q(S_{t+1},a|\theta_{t}^2, g)
\end{equation}

\section{METHODOLOGY}

\subsection{Scenario and Problem Description}

In this paper, we chose the urban scenario of lane change with added complexity. The high-level overview of the scenario can be seen in Figure 1. The ego-car (green car) starts in the ego-lane (lane that has the ego-car) and is required to do a lane change, as there is an obstacle (blue car) in front of it. But the target-lane (the lane where the ego-car chooses to go) has dynamic flow of traffic. In our case, we are using two cars, colored in yellow, to represent the surrounding traffic. This scenario can be easily broken down into tasks with multiple sub-goals, which explains our reason for choosing this scenario. In this problem we are providing our car with two high-level options for the maneuver. The first option is lane follow/wait and the second option is the lane change maneuver. The high-level structure of the proposed HRL trajectory planner can be seen in Figure 1.

\subsection{HRL-based Trajectory Planner}

The division through HRL into two high-level options (lane follow/wait and lane change) helps the ego-car to learn a policy for both high-level options and for the low-level trajectory planners. The details of the hierarchical network structure can be seen in Figure 2.

Once the high-level option is selected, the low-level trajectory planner selects the final waypoint through the network policy. The selection is based on the ego-car's state information through the epsilon-greedy strategy. Once the final waypoint is selected, the ego-car's target speed is calculated using the maximum acceleration or deceleration it can have in order to ensure a smooth sub-trajectory. Then the target speed and final waypoint values are given to the PID controller, which in turn generates longitudinal and lateral control. These sub-trajectories altogether form a complete trajectory, which includes lane follow and lane change maneuvers. The detailed working of the HRL-based trajectory planner, including training details, is shown in Algorithm 1.  The details of the decision-making strategy for low-level planner choices can be found in the sections below:

\begin{figure}[htbp]
    \centering
    \includegraphics[width=0.7\linewidth]{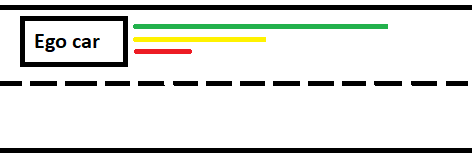}
    \caption{Overview of the lane follow/wait option}
    \label{fig:Waiting}
\end{figure}

\subsubsection{Lane Follow/Wait}

Once the ego-car selects the lane follow/wait option, the low-level trajectory planner is used to plan the path. The low-level planner has three choices for the trajectory, as shown by the colored lines in Figure 3. The green line represents the choice in which the ego-car can follow the lane for a longer distance. This choice is rewarded when the ego-car does not observe any obstacle within a certain radius of its current position in the ego-lane. The yellow line represents a trajectory choice under which the ego-car can follow the lane for a shorter distance. This choice is beneficial when the ego-car observes an obstacle car in the ego-lane at some distance and wants to avoid getting close to it. On the other hand, if the red line trajectory is chosen by the ego-car, it will decelerate to the wait option or a slower speed. Once the trajectory selection is made, the PID controller will be used to complete the path planned.

\begin{figure}[htbp]
    \centering
    \includegraphics[width=0.7\linewidth]{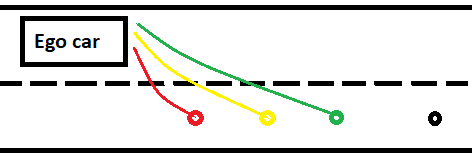}
    \caption{Overview of lane change option}
    \label{fig:Waiting}
\end{figure}

\subsubsection{Lane Change}

Once the decision to perform a lane change is made, the waypoint on the target-lane is selected using the ego-car's state information through the epsilon-greedy strategy. Each different final waypoint shown in Figure 4 represents a different velocity profile the ego-car may choose to make a lane change. This ensures smoothness and stability while planning for the lane change maneuver. The green point is selected if the ego-car is required to do a faster lane change because of the higher speed in the ego-lane. The yellow point is for the normal velocity profile trajectory. The red point selection helps the ego-car to take sharper turns, which might be needed when the velocity of the ego-car is small or it was in the wait option. The RL selects one point between the three points through the low-level planner network under the lane change high-level option. Once the policy selects the point, a black point is generated at some distance in the target-lane. This black point is a safety follow point, which helps the ego-car to reorient itself before the lane follow/wait option is activated again. The generation of the black point is based on the ego-car's distance to the front vehicle after performing lane change.

\subsection{Dealing with Noisy Observations}

In order for our model to consider noisy observations, we added an LSTM layer to both networks: the high-level options and low-level planner network. We used three time steps as an input to the LSTM layer. In order to sample experiences from replay memory, we used the approach of bootstrapped random updates proposed by [19]. This strategy randomly samples an n-step sequence from the batch of episodes drawn from experience replay and then trains the neural network. The hierarchical network structure details can be found in Figure 2.

\subsection{State Space}

In order to formulate the state space for our hierarchical structure, we used information from the ego-car $(e)$, obstacle car $(o)$, target car A $(a)$, and target car B $(b)$ in the given scenario. This can be seen in the tuple given below. The safe range for the ego-car is $x >= 16$ for the stopped or obstacle vehicle and $x >= 9$ for the moving vehicles, where $x$ is the ego-car distance to other vehicles in meters. The state space consists of 14 parameters. The tuple contains the state information where $f$ $\in$ $\{o, a, b\}$: 

$$
s = [ v_e, lane_{ide}, v_f, d_{cf}, d_{cfr}, lane_{idf}]
$$

\begin{itemize}
\item $v_{e}$   = Ego-car velocity
\item $lane_{ide}$ = Lane-ID for the ego-car
\item $v_{f}$   = Velocities of the obstacle car and target vehicles
\item $d_{cf}$ = Ego-car chase distance to the obstacle car in the ego-lane and moving vehicles in the target-lane
\begin{algorithm}[H]
    \caption{HRL-based Trajectory Planner}
    \begin{algorithmic}[1]
        \STATE Initialize options and planner network $Q^{o}$, $Q^{p}$ with weights ${\theta^{o}}$, ${\theta^{p}}$.
        \STATE Initialize target options and target planner network ${Q^{o'}}$, ${Q^{p'}}$ with weights ${\theta^{o'}}$, ${\theta^{P'}}$. 
        \STATE Construct empty replay buffer B with max memory length ${l_{b}}$.
        \STATE Run 100 rule-based episodes and store experience in buffer B.
        \FOR{101 to E training episodes}
            \STATE Get Initial State $s$.
            \STATE Initialize history vector $h_{t}$ containing 3 time steps.
            \WHILE{s is not terminal state}
                \STATE $o_{t}$ = $\argmax_{o}$ $Q(h_{t})$ based on the $\epsilon-greedy$, where $o_{t}$ is the option.
                \STATE $p_{t}$ = $\argmax_{p}$ $Q(h_{t},o_{t})$ based on the $\epsilon-greedy$, where $p_{t}$ is planner waypoint choice.
                \STATE Get the target waypoint $w_{t}$ based on goal $o_{t}$ and planner $p_{t}$ selection.
                \STATE Calculate the target velocity value $v$ based on $w_{t}$.
                \STATE Throttle  = $PID_{longitudinal}(u, v)$, where $u$ is the current velocity and $v$ is the final velocity.
                \STATE Steer = $PID_{lateral}(w_{c},w_{t})$, where $w_{c}$ is the current waypoint and $w_{t}$ is the target waypoint. 
                \STATE Perform the step for sub-trajectory in simulation and get ${s_{t+1}}$, $r^{o}_{t+1}$, $r^{p}_{t+1}$, where $r^{o}$ is the option reward and $r^{p}$ is the planner reward.
                \STATE Deque ${s_{t+1}}$ to $h_{t}$ to get $h_{t+1}$.
                \STATE Store transition $T$ into B: $T$ = ${\{s_{t}, h_{t}, o_{t}, p_{t}, r^{o}_{t+1}, r^{p}_{t+1}, {s_{t+1}}, h_{t+1} \}}$.
            \ENDWHILE
            \STATE Train with Buffer $RelayBuffer(e)$.
        \ENDFOR
    \end{algorithmic}
\end{algorithm}
\item $d_{cfr}$ = Ratio of chase distance to safety threshold
\item $lane_{idf}$ = Lane-ID of the obstacle car and target vehicles
\end{itemize}

\subsection{Reward Structure}

The reward structure consists of a separate high-level options reward and low-level trajectory choice reward. At each time step the following penalties and rewards are given considering the ego-car (e), obstacle car (o), moving car A (a), and moving car B (b):

\begin{itemize}

\item Regular time step penalty: $-\sigma_{1}$
\item Regular time step reward for progressing towards final destination: $\sigma_{2}$
\item Collision penalty: $-\sigma_{3}$
\item Unsafe penalty: $\exp{-(d_{cfr})}$, where $f$ $\in$ $\{o, a, b\}$
\item Goal not required penalty: $-\sigma_{4}$
\item Unsmoothness penalty $-\sigma_{5}$
\item Success Reward $\sigma_{6}$
\end{itemize}
\begin{table*}
\centering
  \caption{Comparison of results obtained from different planning algorithms}
  \label{tab:commands}
  \begin{tabular}[5pt]{|c||c|c|c|c|c|}
    \hline
    \textbf{Method} & \textbf{Gaussian Noise } & \textbf{Total Average Reward} & \textbf{Lane Invasion Rate \%} & \textbf{Collision Rate \%}  & \textbf{Success Rate \%}\\
    \hline
    DDQN + PID & No & 3623.45 & 0 & 8 & 92\\
    \hline
    Hierarchical DDQN (hDDQN) & No & 4245.65 & 15 & 4 & 96\\
    \hline
    Slot-based + PID & No & 4510.70 & 0 & 1 & 99\\
    \hline
    Slot-based + PID & Yes & 3200.82 & 0 & 19 & 81\\
    \hline
    hDDQN + PID & No & 4675.64 & 0 & 0 & 100\\
    \hline
    hDDQN + PID & Yes & 3951.00 & 0 & 7 & 93\\
    \hline
    \textbf{hDDQN + PID + LSTM} & \textbf{No} & \textbf{4615.38} & \textbf{0} & \textbf{0} & \textbf{100}\\
    \hline
    \textbf{hDDQN + PID + LSTM} & \textbf{Yes} & \textbf{4451.01} & \textbf{0} & \textbf{2} & \textbf{98}\\
    \hline
  \end{tabular}
\end{table*} 
The ego-car gets penalized separately for choosing the wrong high-level option or choosing the wrong low-level trajectory choice. The low-level choice gets penalized if the chosen trajectory leads to unsuccessful completion of the sub-goal, which in our case means collision with one of the target cars or the obstacle car. Moreover, in order to plan safer and smoother trajectories, low-level choices were penalized if they were not required. For example, under option 1, the low-level wait choice gets penalized if it is selected unnecessarily. The ego-car is penalized more as the safe chase distance to other vehicles decreases. The safe distance is the minimum distance that the ego-car has to maintain from the obstacle car or from the moving vehicles. The ego-car is expected to plan a longer trajectory if it does not see any obstacle in the ego-lane within a certain distance.
For the lane change maneuver, the ego-car gets a reward for the safest point selected based on its past velocity profile. For example, if the ego-car has a higher velocity in the target-lane, it is given a higher reward to do a faster lane change; otherwise, if the ego-car was waiting in the ego-lane, it is expected to do a slower turn. This summarizes the unsmoothness penalty. These different low-level point choices prevent lane invasion and unnecessary jerk.

\section{Experiments}

\subsection{Scenario and Experiment Setup}

In this section, we apply the proposed HRL algorithm to the scenario shown in Figure 1 in the CARLA simulator [20] with 30 fps. The situation is similar to the real-world scenario when there are parked cars in the ego-lane and a traffic flow in the target-lane. In this situation, the ego-car is expected to plan trajectories for either lane follow or lane change maneuvers considering both the moving and stopped vehicles. The ego-car has to avoid collision with the parked car in the ego-lane and the traffic in the target-lane. It also has to make sure that sub-trajectory choices constitute an overall smooth trajectory. In order to test the generality and safety of our approach, we made surroundings of the ego-car more dynamic during the training stage in the following two ways: randomly changing the 1) target cars' velocities and 2) the position of the obstacle car.

Moreover, to generate a better experience replay buffer for training, we ran the starting 100 episodes using the high-level rule-based methods, which allowed the ego-car to discern when to select which option. The epsilon-greedy strategy-based training started after 100 episodes. Furthermore, in order to evaluate the performance of our algorithm with inaccurate or incomplete observations, we added Gaussian noise to the input space of the options and planner network. Algorithm 1 describes the proposed Hierarchical RL algorithm. Lastly, in order to evaluate the effectiveness of our proposed algorithm, we compared it with five other methods: 1) Double Deep Q-Learning (DDQN) with PID, 2) Hierarchical Double Q-Learning with no PID, 3) Slot-based method with PID controller, 4) Hierarchical Double Q-Learning with PID, and 5) Hierarchical Double Q-Learning with PID and an LSTM layer. The slot-based method consisted of rules for the high-level options and low-level choices. For each method, the performance was recorded, both with and without Gaussian noise. Furthermore, for each method, training consisted of 2000 episodes and the policy evaluation after training consisted of 200 episodes. We compared the performance of different approaches using these metrics:
\begin{itemize}
\item \textbf{Total Average Reward}: The summation of high-level option reward and the low-level planner choice reward divided by total number of test episodes;
\item \textbf{Lane Invasion rate}: The average rate of lane invasion recorded in the test episodes. Lane invasion occurs when the ego-car crosses the ego-lane's boundaries while in the following lane state. Lane invasion usually occurs due to poor control of the vehicle or unsmooth trajectory. This value does not include the lane invasion instance when the ego-car performs a lane change;
\item \textbf{Collision rate}: The percentage of test episodes in which collision occurs;
\item \textbf{Success rate}: The percentage of test episodes in which the ego-car is able to complete its trajectory from starting point to the end point without collision.
\end{itemize}

\subsection{Results and Discussion}

The performance of several methods was evaluated and the detailed results are shown in Table I. The training results for the some of the approaches used is shown in the Figure 5. For the proposed methodology, we were able to get the optimal policy after 2000 episodes. The proposed hierarchical structure recorded 100\% success rate without noisy observations and 98\% success rate with noisy state space.

\begin{figure}[htbp]
    \centering
    \includegraphics[width=1.0\linewidth]{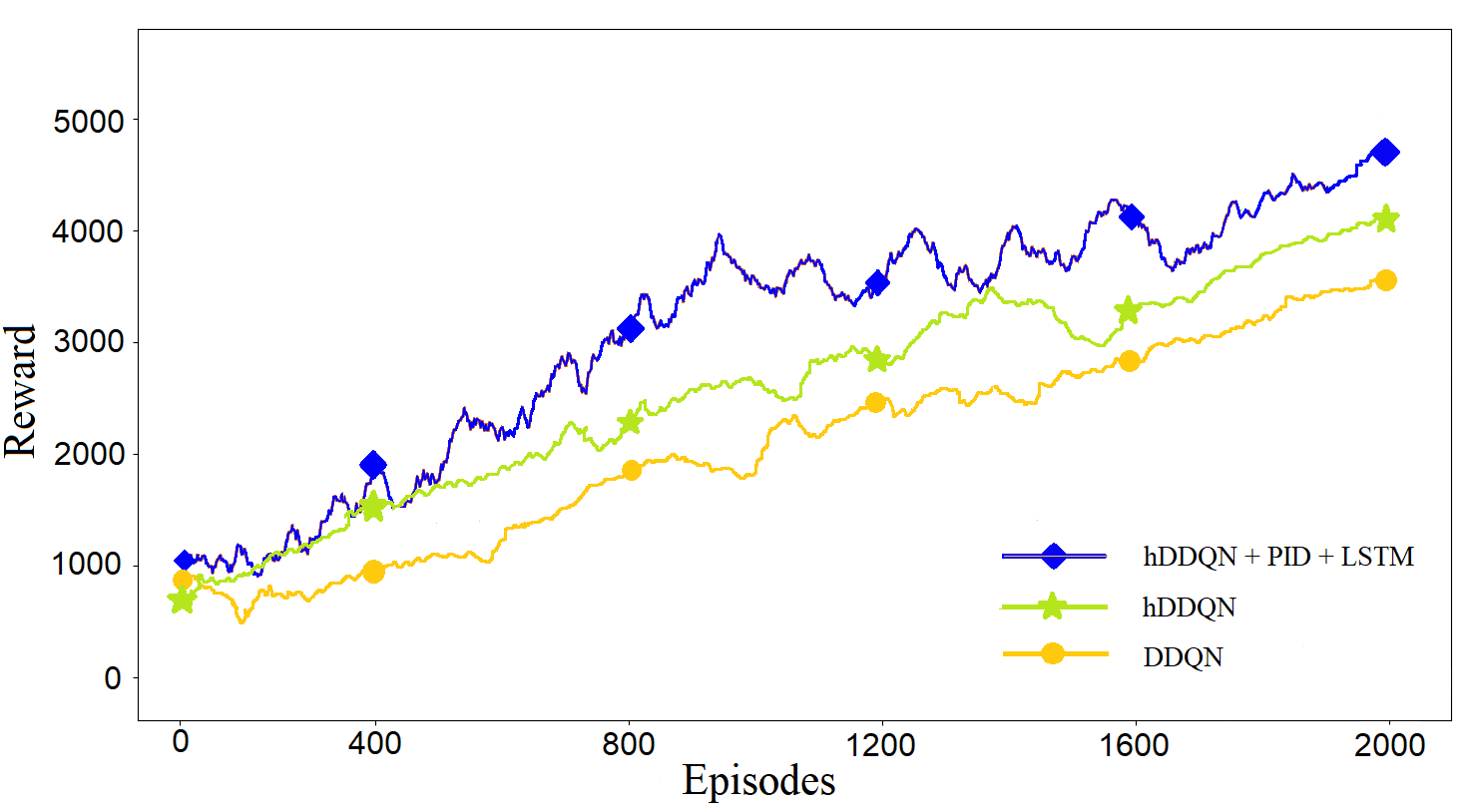}
    \caption{Training results of different approaches}
    \label{fig:Waiting}
\end{figure}

For the convergence part, direct comparisons with non-hierarchical approaches suggest that the presence of sub-goals through the use of HRL decreased the overall convergence time. The hierarchical structure also allowed us to design the reward structure for high-level options and low-level choices separately, which is one of the
reasons for the better performance of HRL. Table I shows that simple DDQN only achieved a success rate of 92\% in comparison to the hierarchical approaches, which were able to have a minimum 96\% success rate with no noise added to the state space and with direct low-level control. Also in comparison to classical planners such as RRT's and learning-based techniques such as Imitation Learning, our learning-based approach can handle dynamic surroundings well and produce smooth trajectories without increasing computational complexity.

The results and evaluation of final policies show that using the PID controller instead of directly using throttle, steer, and brake ensured smooth tracking and increased safety. As the waypoints were generated in the middle of the road by the CARLA waypoint API, the ego-car recorded almost zero lane invasions while tracking the selected waypoints using the PID controller. Tracking of waypoints ensures safety and complete control of the ego-car throughout the trajectory as compared to classical planners such as lattice planners, which might produce incomplete graphs. In addition, as the velocity of the ego-car was selected based on target waypoint and maximum acceleration or deceleration value, the jerk was minimized.

In order to evaluate the performance with incomplete observations, we added Gaussian noise to the state space. For most of the methods, we trained the model with and without noise. This helped us to make direct comparisons with other approaches. The results in Table I show that the collision rate in the hDDQN with PID was brought down to 2\% from 7\% with the inclusion of the LSTM layers in the network. Although improved, the collision rate was not able to drop to zero after 2000 episodes of training, as LSTMs require more time to train. But overall, adding an LSTM layer to the network yielded the best results with the noisy state space after 2000 episodes of training. 

\section{Conclusion}

In this paper, we propose a trajectory planning algorithm for autonomous vehicles based on Hierarchical Reinforcement Learning (HRL). Through this structure, the ego-car is given high-level goal options and low-level planner choices. A PID controller is used instead of directly using throttle, steering, and brake. The results show that the proposed framework decreases convergence time, ensures safe and smooth tracking of waypoints using PID, and is able to handle dynamic surroundings and noisy observations by using an LSTM layer for both networks in the hierarchical structure. The future work includes combining policies learned from different scenarios and maneuvers through HRL as sub-goals. Moreover, HRL-based trajectory planner algorithm performance can be compared by using controllers other than PID such as Model Predictive Controller (MPC).


\section*{ACKNOWLEDGMENT}

The authors would like to thank the Robotics Institute Summer Scholars Program, Carnegie Mellon University, USA and The Industrial Centre, The Hong Kong Polytechnic University, Hong Kong for the support and providing resources for remote research.


\nocite{*}  
\bibliographystyle{IEEEtran}
\bibliography{./IEEEfull,refs}

\begin{thebibliography}{10}
\providecommand{\url}[1]{#1}
\csname url@rmstyle\endcsname
\providecommand{\newblock}{\relax}
\providecommand{\bibinfo}[2]{#2}
\providecommand\BIBentrySTDinterwordspacing{\spaceskip=0pt\relax}
\providecommand\BIBentryALTinterwordstretchfactor{4}
\providecommand\BIBentryALTinterwordspacing{\spaceskip=\fontdimen2\font plus
\BIBentryALTinterwordstretchfactor\fontdimen3\font minus
  \fontdimen4\font\relax}
\providecommand\BIBforeignlanguage[2]{{%
\expandafter\ifx\csname l@#1\endcsname\relax
\typeout{** WARNING: IEEEtran.bst: No hyphenation pattern has been}%
\typeout{** loaded for the language `#1'. Using the pattern for}%
\typeout{** the default language instead.}%
\else
\language=\csname l@#1\endcsname
\fi
#2}}

\bibitem{c1}
S.~LaValle, J.~Kuffner, and B.~Donaldetal, ``Rapidly-exploring random trees:
  Progress and prospects,'' in \emph{Algorithmic and computational robotics:
  new directions}, 2001, pp. 293--308.

\bibitem{c2}
H.~Mouhagir, R.~Talj, V.~Cherfaoui, F.~Guillemard, and F.~Aioun, ``A markov
  decision process-based approach for trajectory planning with clothoid
  tentacles,'' in \emph{2016 IEEE Intelligent Vehicles Symposium (IV)}, 2016,
  pp. 1254--1259.

\bibitem{c3}
W.~Xu, J.~Wei, J.~M. Dolan, H.~Zhao, and H.~Zha, ``A real-time motion planner
  with trajectory optimization for autonomous vehicles,'' in \emph{2012 IEEE
  International Conference on Robotics and Automation}, 2012, pp. 2061--2067.

\bibitem{c4}
P.~{Cai}, Y.~{Sun}, Y.~{Chen}, and M.~{Liu}, ``Vision-based trajectory planning
  via imitation learning for autonomous vehicles,'' in \emph{2019 IEEE
  Intelligent Transportation Systems Conference (ITSC)}, 2019, pp. 2736--2742.

\bibitem{c5}
T.~Osa, J.~Pajarinen, G.~Neumann, J.~Bagnell, P.~Abbeel, and J.~Peters, ``An
  algorithmic perspective on imitation learning,'' \emph{Foundations and Trends
  in Robotics}, vol.~7, no. 1-2, pp. 1--179, Mar. 2018.

\bibitem{c6}
R.~S. Sutton and A.~G. Barto, \emph{Reinforcement Learning: An
  Introduction.}\hskip 1em plus 0.5em minus 0.4em\relax Cambridge, MA, USA: MIT
  Press, 1998.

\bibitem{c7}
T.~D. Kulkarni, K.~Narasimhan, A.~Saeedi, and J.~Tenenbaum, ``Hierarchical deep
  reinforcement learning: Integrating temporal abstraction and intrinsic
  motivation,'' in \emph{Advances in Neural Information Processing Systems 29},
  2016, pp. 3675--3683.

\bibitem{c8}
T.~G. Dietterich, ``The maxq method for hierarchical reinforcement learning.''
  \emph{ICML}, vol.~98, pp. 1--179, 1998.

\bibitem{c9}
N.~K. Jong and P.~Stone, ``Hierarchical model-based reinforcement learning:
  R-max+ maxq,'' \emph{in Proceedings of the 25th international conference on
  Machine learning. ACM}, pp. 1--179, 2008.

\bibitem{c10}
Z.~{Qiao}, K.~{Muelling}, J.~{Dolan}, P.~{Palanisamy}, and P.~{Mudalige},
  ``Pomdp and hierarchical options mdp with continuous actions for autonomous
  driving at intersections,'' in \emph{2018 21st International Conference on
  Intelligent Transportation Systems (ITSC)}, 2018, pp. 2377--2382.

\bibitem{c11}
Z.~Qiao, Z.~Tyree, P.~Mudalige, J.~Schneider, and J.~Dolan, ``Hierarchical
  reinforcement learning method for autonomous vehicle behavior planning,''
  Nov. 2019.

\bibitem{c12}
S.~Karaman and E.~Frazzoli, ``Sampling-based algorithms for optimal motion
  planning,'' \emph{The International Journal of Robotics Research}, vol.~30,
  no.~7, pp. 846--894, 2011.

\bibitem{c13}
J.~Chen, B.~Yuan, and M.~Tomizuka, ``Deep imitation learning for autonomous
  driving in generic urban scenarios with enhanced safety,'' 2019.

\bibitem{c14}
Z.~Qiao, J.~Zhao, Z.~Tyree, P.~Mudalige, J.~Schneider, and J.~Dolan, ``Human
  driver behavior prediction based on urbanflow,'' 2019.

\bibitem{c15}
Y.~Zhang, W.~Wang, R.~Bonatti, D.~Maturana, and S.~Scherer, ``Integrating
  kinematics and environment context into deep inverse reinforcement learning
  for predicting off-road vehicle trajectories,'' \emph{CoRR}, vol.
  abs/1810.07225, 2018.

\bibitem{c16}
M.~L. Ho, P.~T. Chan, and A.~B. Rad, ``Lane change algorithm for autonomous
  vehicles via virtual curvature method,'' \emph{Journal of Advanced
  Transportation}, vol.~43, no.~1, pp. 47--70, 2009.

\bibitem{C17}
\BIBentryALTinterwordspacing
T.~Shi, P.~Wang, X.~Cheng, and C.~Y. Chan, ``Driving decision and control for
  autonomous lane change based on deep reinforcement learning,'' \emph{CoRR},
  vol. abs/1904.10171, 2019. [Online]. Available:
  \url{http://arxiv.org/abs/1904.10171}
\BIBentrySTDinterwordspacing

\bibitem{c18}
\BIBentryALTinterwordspacing
C.~J. Hoel, K.~Wolff, and L.~Laine, ``Automated speed and lane change decision
  making using deep reinforcement learning,'' \emph{CoRR}, vol. abs/1803.10056,
  2018. [Online]. Available: \url{http://arxiv.org/abs/1803.10056}
\BIBentrySTDinterwordspacing

\bibitem{c19}
M.~Hausknecht and P.~Stone, ``Deep recurrent q-learning for partially
  observable mdps,'' 2015.

\bibitem{c20}
\BIBentryALTinterwordspacing
A.~Dosovitskiy, G.~Ros, F.~Codevilla, A.~M. L{\'{o}}pez, and V.~Koltun,
  ``{CARLA:} an open urban driving simulator,'' \emph{CoRR}, vol.
  abs/1711.03938, 2017. [Online]. Available:
  \url{http://arxiv.org/abs/1711.03938}
\BIBentrySTDinterwordspacing

\end{thebibliography}

\end{document}